\pdfoutput=1
\documentclass[10pt,twocolumn,letterpaper]{article}

\usepackage{cvpr}
\usepackage{times}
\usepackage{epsfig}
\usepackage{graphicx}
\usepackage{amsmath}
\usepackage{amssymb}

\usepackage[pagebackref=true,breaklinks=true,letterpaper=true,colorlinks,bookmarks=false]{hyperref}

\cvprfinalcopy 

\def\cvprPaperID{****} 

\ifcvprfinal\fi
\begin{document}

\title{1-HKUST: Object Detection in ILSVRC 2014\thanks{indicates equal contribution,
all the authors are united under {\bf 1-HKUST} although some of their current
affiliations are elsewhere.} }

\author{
\begin{tabular}{cccccc}
Cewu Lu$^\dagger$ &
Hao Chen$^{*\ddagger}$ &
Qifeng Chen$^{*\S}$ &
Hei Law$^{*\dagger}$ &
Yao Xiao$^{*\dagger}$ &
Chi-Keung Tang$^\dagger$
\end{tabular} \\
$^\dagger$Hong Kong University of Science and Technology\\
$^\ddagger$The Chinese University of Hong Kong\\
$^\S$Stanford University\\
}

\maketitle

\begin{abstract}
The Imagenet Large Scale Visual Recognition Challenge (ILSVRC) is the one of the most important big data challenges to date. We participated in the object detection track of ILSVRC 2014 and received the fourth place among the 38 teams.  We introduce in our object detection system a number of novel techniques in localization and recognition.  For localization, initial candidate proposals are generated using selective search, and a novel bounding boxes regression method is used for better object localization. For recognition, to represent a candidate proposal, we adopt three features, namely, RCNN feature, IFV feature, and DPM feature.  Given these features, category-specific combination functions are learned to improve the object recognition accuracy.  In addition, object context in the form of background priors and object interaction priors are learned and applied in our system. Our ILSVRC 2014 results are reported alongside with the results of other participating teams.

\end{abstract}



\section{Introduction}\label{sec:intro}
We present our system for the ILSVRC 2014 competition. The big data challenge has evolved
over the past few years as one for the most important forums for researchers to exchange
their ideas, benchmark their systems, and push breakthrough in categorical
object recognition in large-scale image classification and object detection.

The 1-HKUST team made its debut in this year's competition and we focused on
the object detection track.  The object detection problem can be divided into
two sub-problems, namely, localization and recognition.  Localization solves
the ``where" problem, while recognition solves the ``what" problem.  That is,
we locate where the objects are, and then recognize which object categories
the detected objects should belong to.

We made technical contributions on both localization and recognition.
For localization, we exploit regression on bounding box using
deep learning based on selective search outputs. For recognition,
we focus on integrating the state-of-the-art computer vision techniques
to build a more powerful category-specific category predictor. In addition,
object background priors are also considered.

\section{Framework}\label{sec:framework}

\begin{figure}[tb]
\begin{center}
\begin{tabular}{@{\hspace{0mm}}c}
\includegraphics[width=1\linewidth]{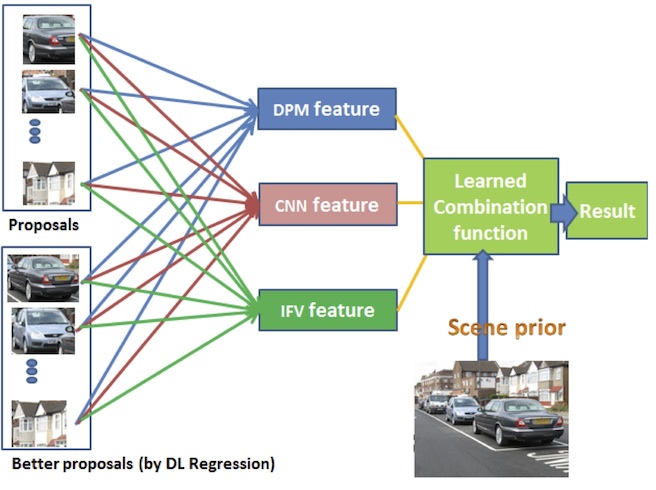}
\end{tabular}
\caption{Our framework.} \label{fig:framework}
\end{center}
\end{figure}

Figure~\ref{fig:framework} gives the overview of our system which
summarizes our contributions in object localization and recognition.

\subsection{Localization}
We first extract candidate objectness proposals using selective search.  As widely
known, the output bounding boxes are almost never perfect and fail to coincide
the ground-truth object boxes with a high overlap rate (e.g. $80\%$).  To cope
with this problem, we learn a regressor using deep learning.

\subsection{Recognition}
Given a set of candidate proposals in hand, we extract different types
of feature representation for recognition.  We adopt three types
of feature, namely, CNN feature, DPM feather and IFV feature, to measure
the given candidate proposals.

For CNN feature, we first train the CNN model similar with CaffeNet (refer to~\cite{Jia13caffe} for architecture details), and the outputs of the Fc6 layer are extracted as the CNN features.  We apply the SVM training to obtain 200 object category classifiers, as similarly done in RCNN~\cite{Girshick2014Rich}.  For DPM feature~\cite{Felzenszwalb2010discriminatively} we also train 200 DPM models.  For IFV feature~\cite{Perronnin2010Improving}, we make use of the fast IFV feature extraction solution~\cite{Koen2014Fisher} to compute at a rate of 20 seconds per image.  We also train 200 SVM category models as similarly done for the above two features.  After obtaining 200 CNN scores, 200 DPM scores, and 200 IFV scores, these scores are concatenated into a 600-dimensional feature vector.  Finally, we train a 200-class SVM model on these features.

\subsection{Background Prior}
Objects occur in context and are part of the scene. Background scene understanding can definitely benefit object detection.  The background can reject (or re-score) unreasonable objects.  For example, a yacht does not appear in an indoor environment with high probability.  In our implementation, we train a presence prior model (PPM) under the CNN framework on the object detection data of ILSVRC 2014.  Rather than producing a single label per image, this method outputs multiple labels for an image.  Thus, false predictions can be removed if the prediction score based on our trained presence prior falls below a confidence threshold.  Our experimental results demonstrate that the presence prior could help to filter false predictions with more context information being considered. 

\section{Results}
\begin{figure*}[tb]
\begin{center}
\begin{tabular}{@{\hspace{0mm}}c@{\hspace{1mm}}c@{\hspace{1mm}}c@{\hspace{1mm}}c@{\hspace{0mm}}c@{\hspace{1mm}}c@{\hspace{1mm}}c@{\hspace{1mm}}c}
\includegraphics[width=0.13\linewidth]{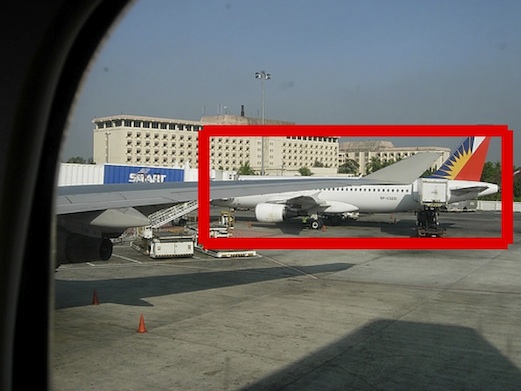}&
\includegraphics[width=0.13\linewidth]{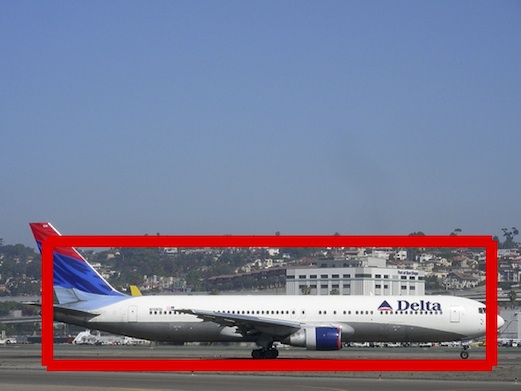}&
\includegraphics[width=0.13\linewidth]{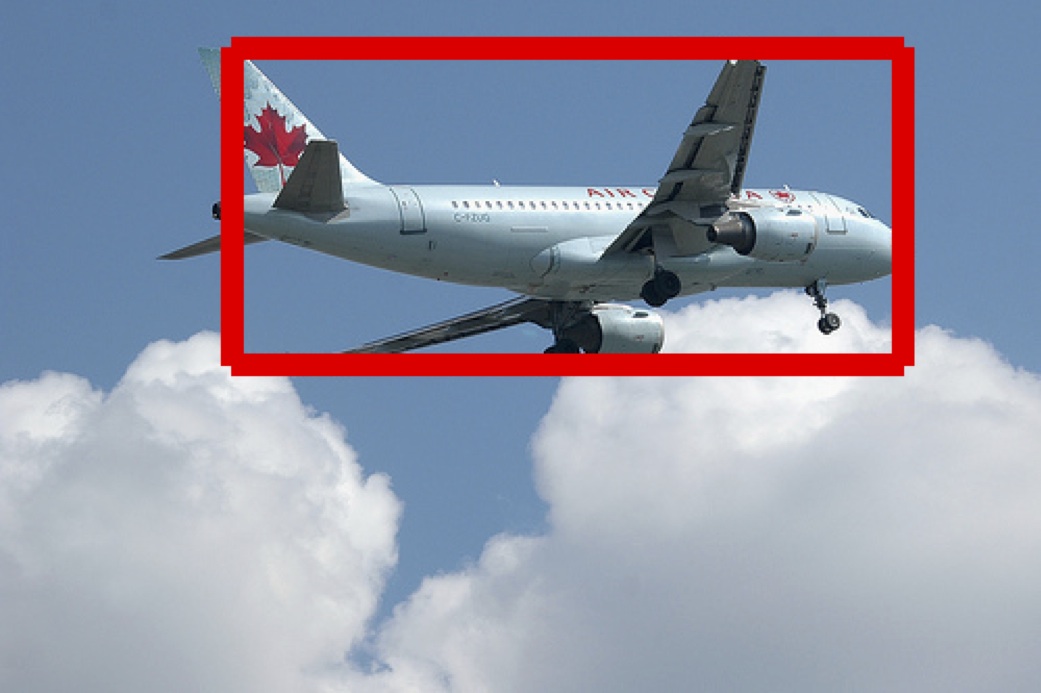}&
\includegraphics[width=0.13\linewidth]{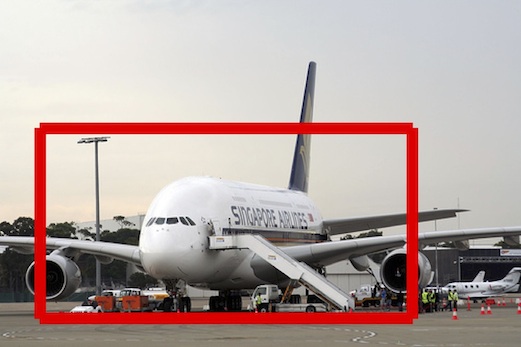}&
\includegraphics[width=0.13\linewidth]{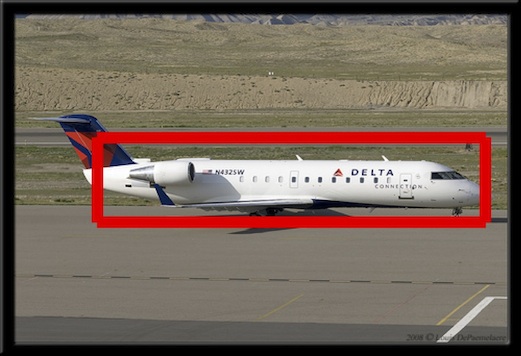}&
\includegraphics[width=0.13\linewidth]{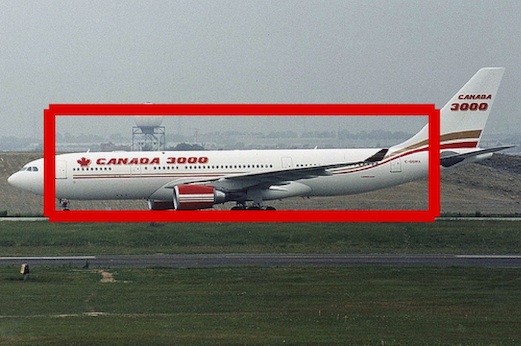}&
\includegraphics[width=0.13\linewidth]{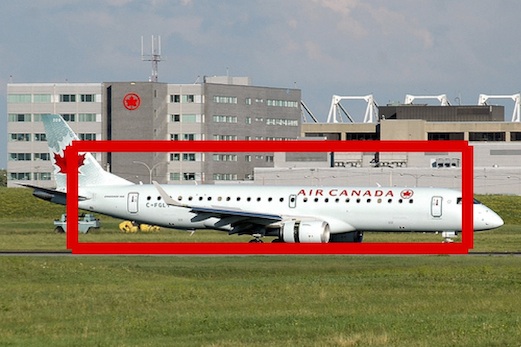}
\\
airplane
\\
\includegraphics[width=0.13\linewidth]{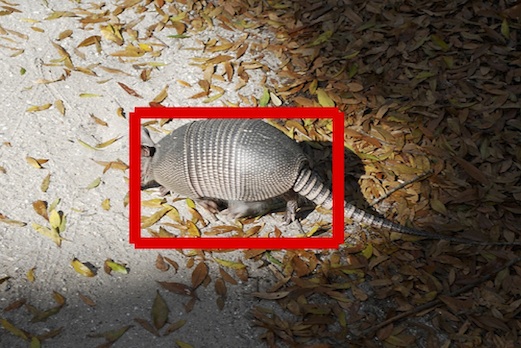}&
\includegraphics[width=0.13\linewidth]{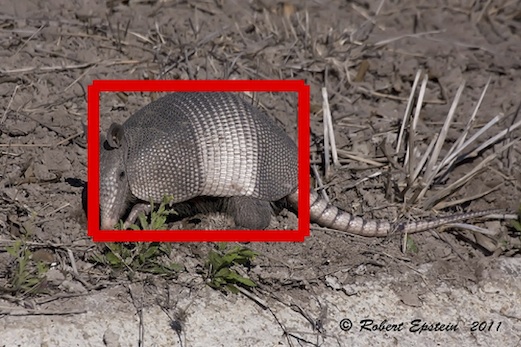}&
\includegraphics[width=0.13\linewidth]{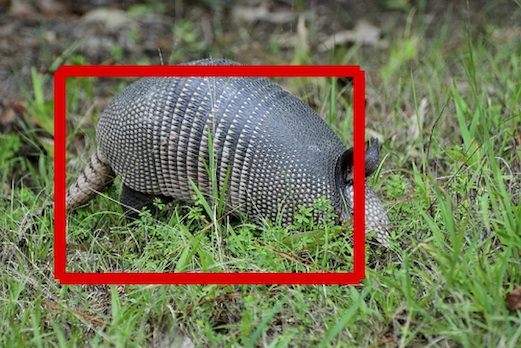}&
\includegraphics[width=0.13\linewidth]{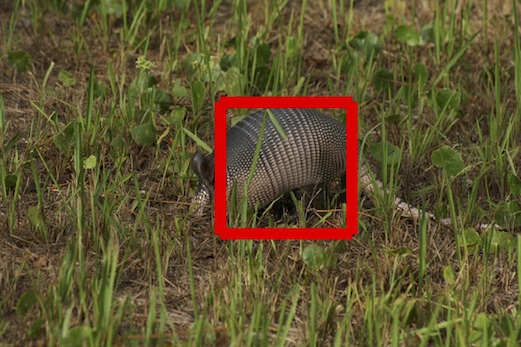}&
\includegraphics[width=0.13\linewidth]{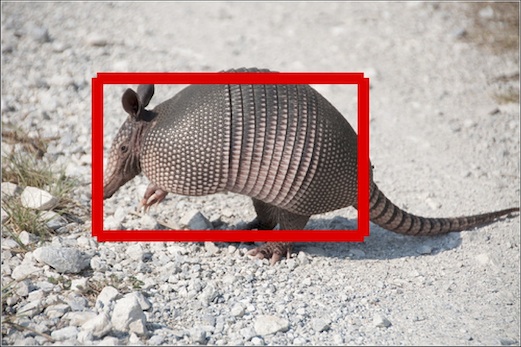}&
\includegraphics[width=0.13\linewidth]{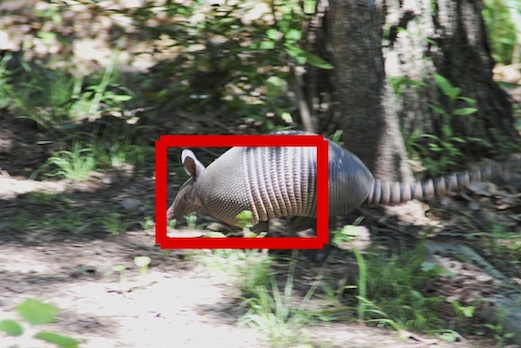}&
\includegraphics[width=0.13\linewidth]{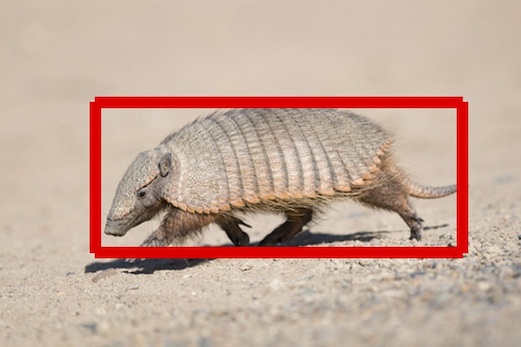}
\\
armadillo
\\
\includegraphics[width=0.13\linewidth]{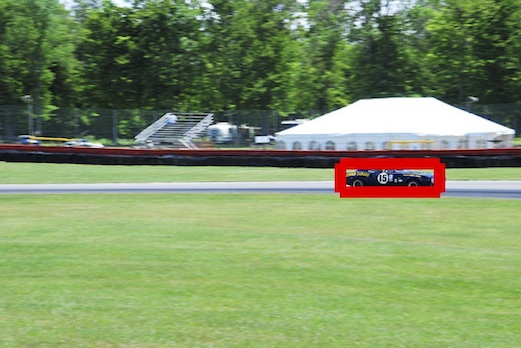}&
\includegraphics[width=0.13\linewidth]{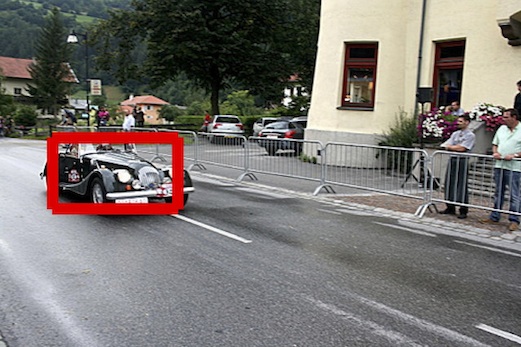}&
\includegraphics[width=0.13\linewidth]{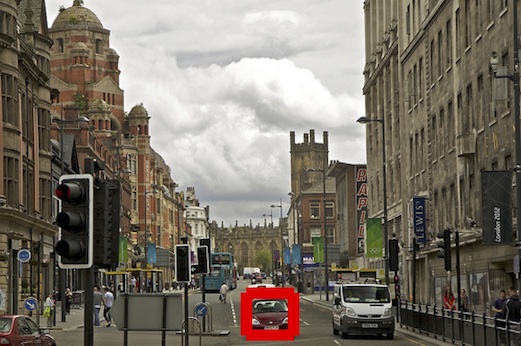}&
\includegraphics[width=0.13\linewidth]{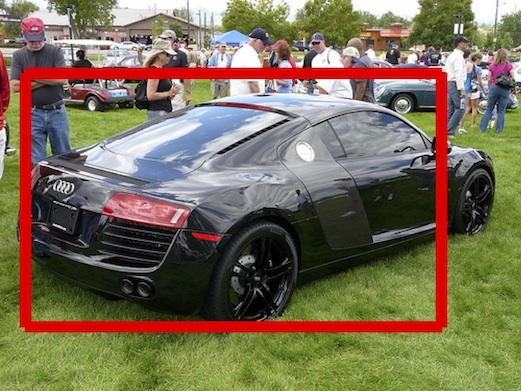}&
\includegraphics[width=0.13\linewidth]{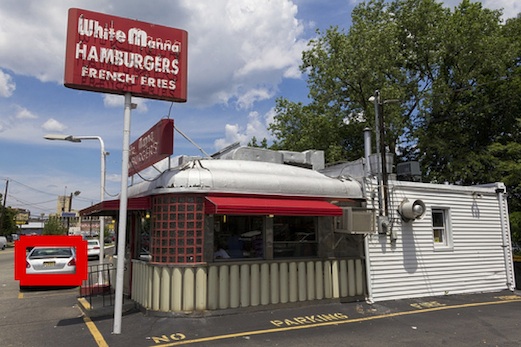}&
\includegraphics[width=0.13\linewidth]{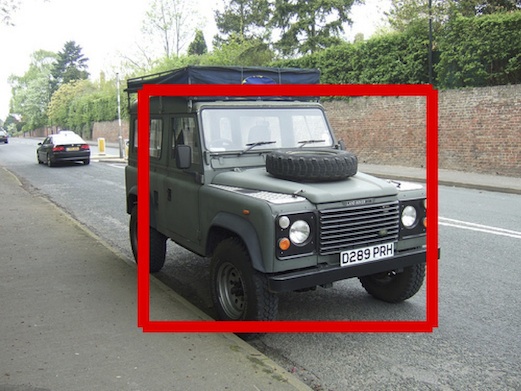}&
\includegraphics[width=0.13\linewidth]{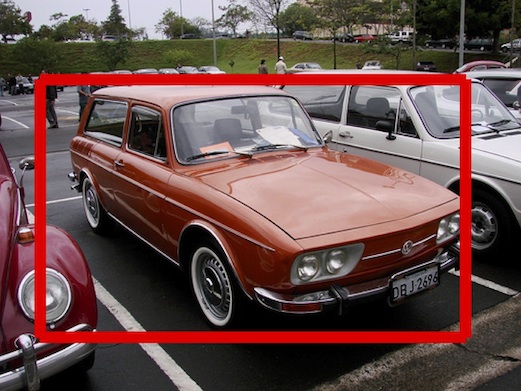}
\\
car
\\
\includegraphics[width=0.13\linewidth]{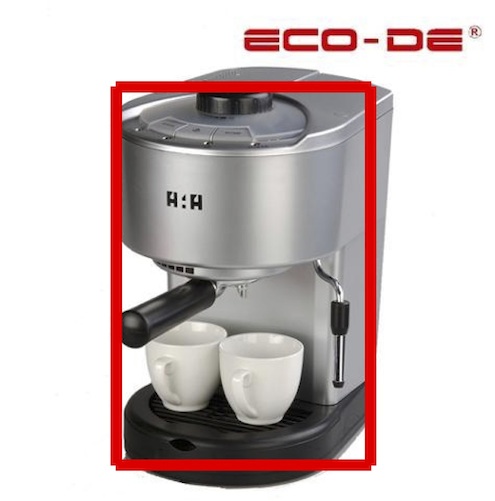}&
\includegraphics[width=0.13\linewidth]{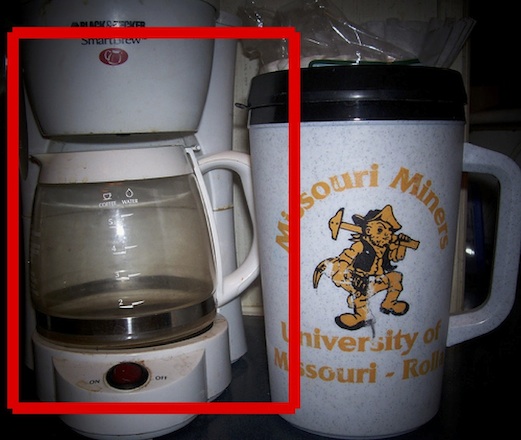}&
\includegraphics[width=0.13\linewidth]{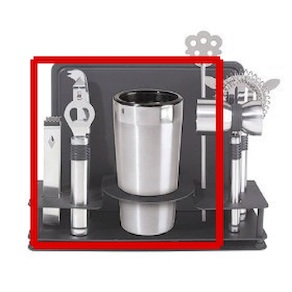}&
\includegraphics[width=0.13\linewidth]{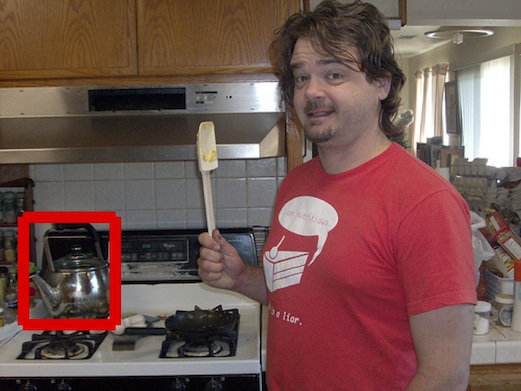}&
\includegraphics[width=0.13\linewidth]{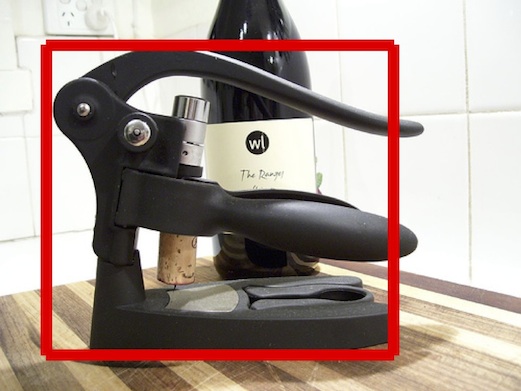}&
\includegraphics[width=0.13\linewidth]{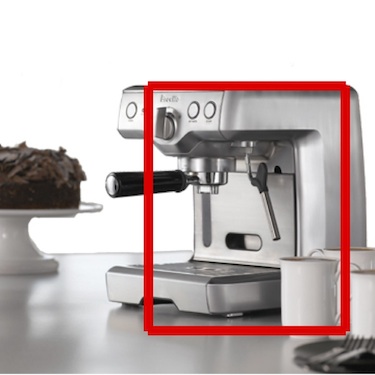}&
\includegraphics[width=0.13\linewidth]{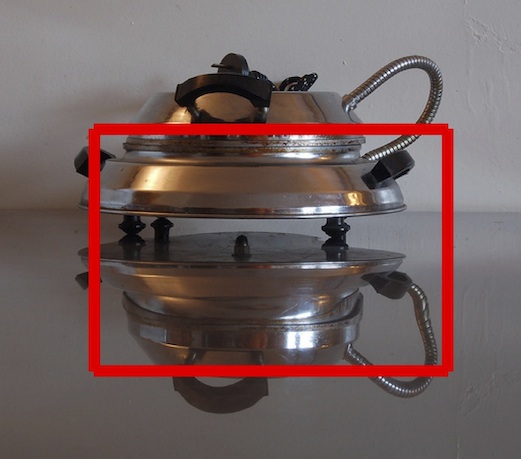}
\\
coffee maker
\\
\includegraphics[width=0.13\linewidth]{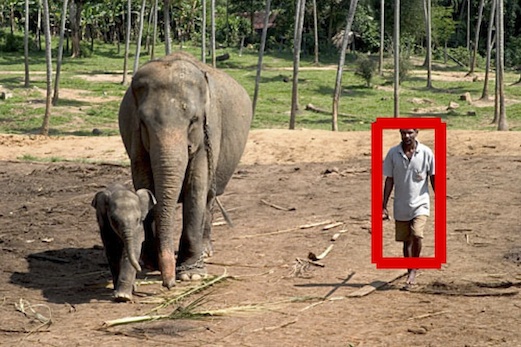}&
\includegraphics[width=0.13\linewidth]{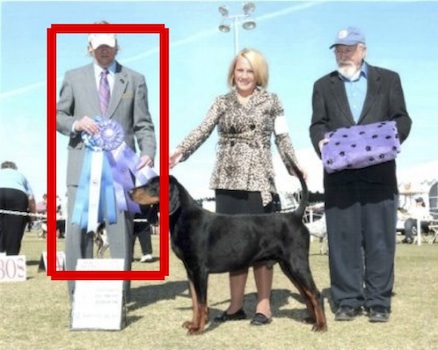}&
\includegraphics[width=0.13\linewidth]{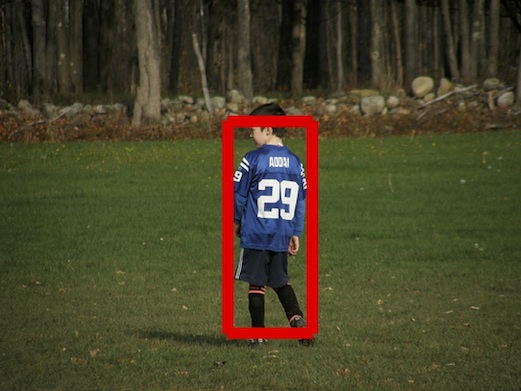}&
\includegraphics[width=0.13\linewidth]{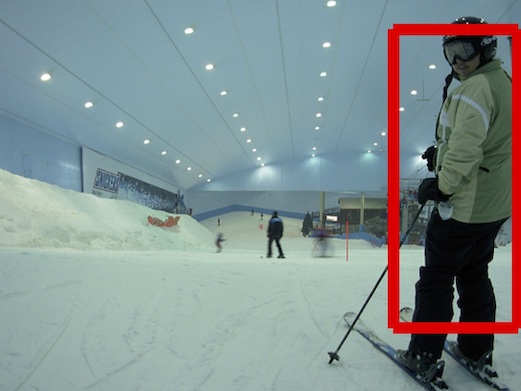}&
\includegraphics[width=0.13\linewidth]{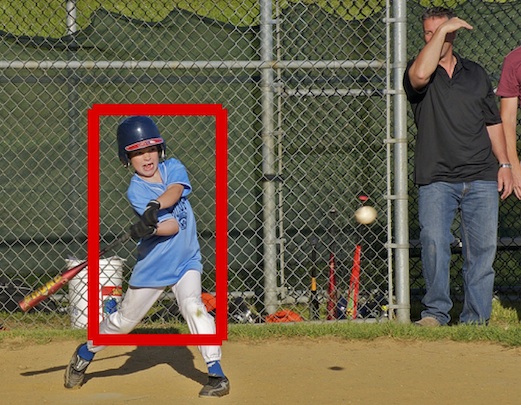}&
\includegraphics[width=0.13\linewidth]{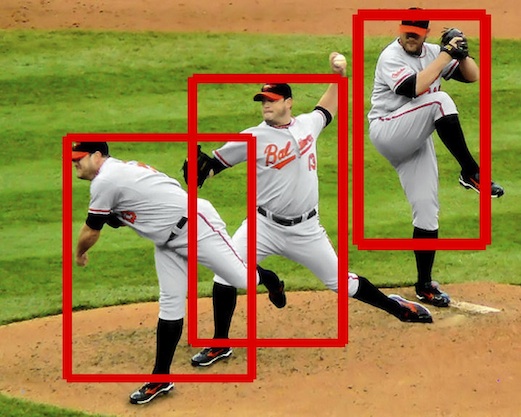}&
\includegraphics[width=0.13\linewidth]{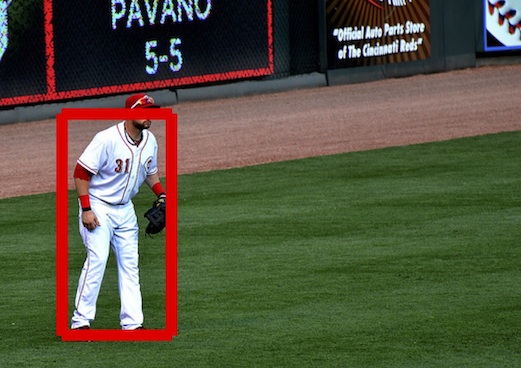}
\\
person
\\
\includegraphics[width=0.13\linewidth]{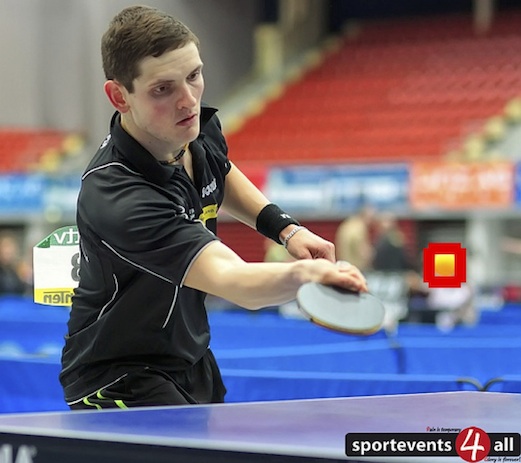}&
\includegraphics[width=0.13\linewidth]{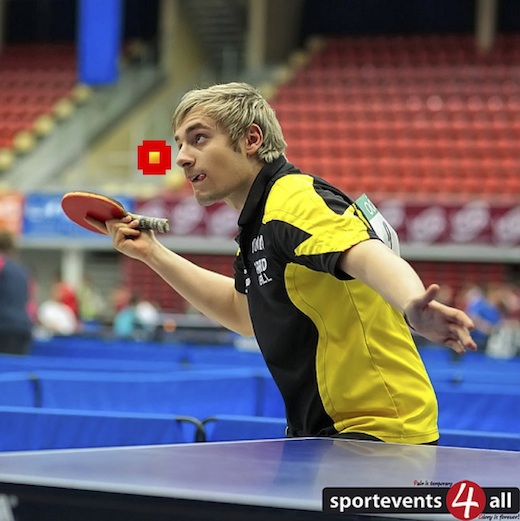}&
\includegraphics[width=0.13\linewidth]{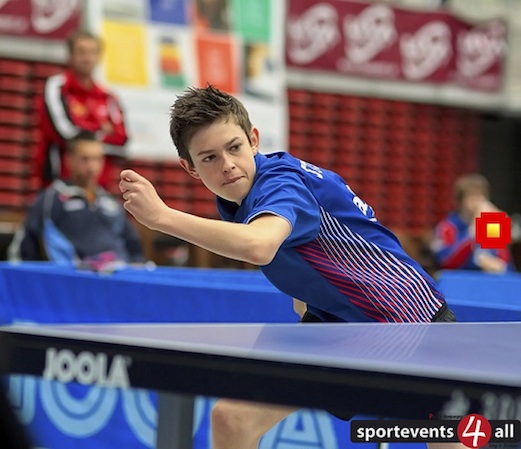}&
\includegraphics[width=0.13\linewidth]{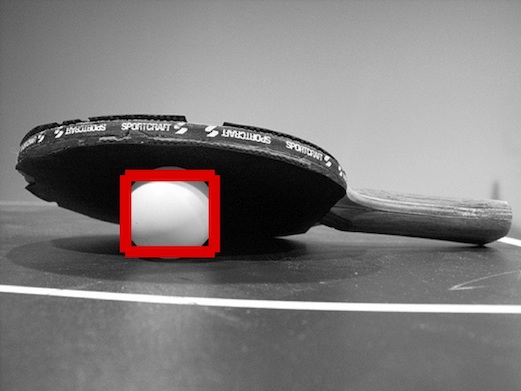}&
\includegraphics[width=0.13\linewidth]{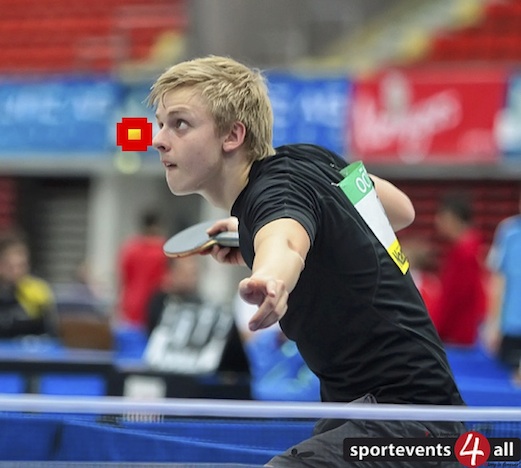}&
\includegraphics[width=0.13\linewidth]{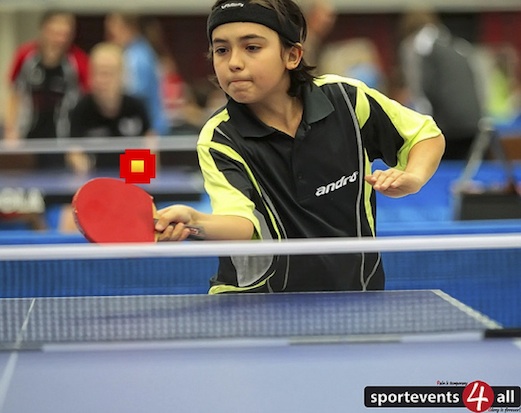}&
\includegraphics[width=0.13\linewidth]{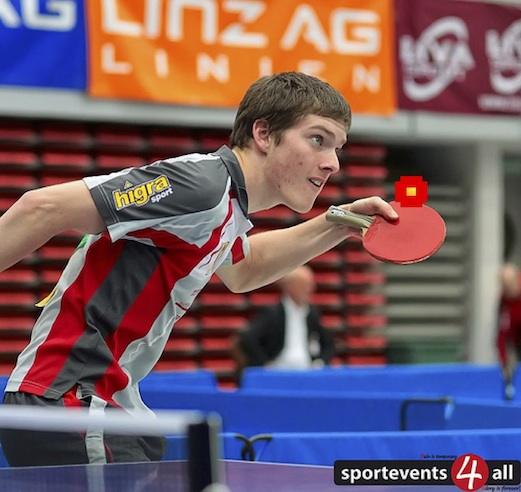}
\\
ping pong
\\
\end{tabular}
 \vspace{2mm}
\caption{Sample object detection results. } \label{fig:result}
\end{center}
\end{figure*}

\begin{figure}[tb]
\begin{center}
\begin{tabular}{@{\hspace{0mm}}c@{\hspace{1mm}}c@{\hspace{1mm}}c@{\hspace{1mm}}c@{\hspace{0mm}}c@{\hspace{1mm
}}c@{\hspace{1mm}}c@{\hspace{1mm}}c}
\includegraphics[width=0.40\linewidth]{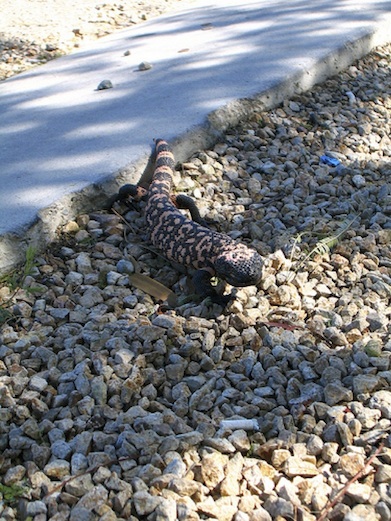}&
\includegraphics[width=0.40\linewidth]{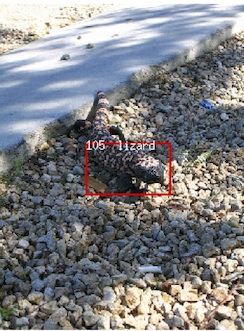}
\\
(a) & (b)
\\
\end{tabular}
 \vspace{2mm}
\caption{(a) is a input image, (b) is our detection result.
Some people found it difficult to recognize a lizard on pebbles.  }
\label{fig:case}
\end{center}
\end{figure}

We discuss the performance of our entries in ILSVRC 2014. In the object detection track,
there are two sub-tracks: with and without extra training data. Our results were achieved
{\em without} extra training data.  We were ranked fourth in terms of number of winning
categories.  Table~\ref{tab:result1} tabulates the top winners and we refer readers to~\cite{ILSVRCarxiv14} or the official website of ILSVRC 2014 for complete standings.  Our mAP is $0.289$. By analyzing the per-class results\footnote{\url{http://image-net.org/challenges/LSVRC/2014/} \\ \url{results/ilsvrc2014_perclass_results.zip}}, we found that 1-HKUST is still ranked fourth among all the teams using and without using extra training data.
Table~\ref{tab:result2} shows that using extra training data gives a clear advantage.
Sample visual results are demonstrated in Figure~\ref{fig:result}. Surprisingly, a number of
difficult cases for human detection such as the lizard in~Figure~\ref{fig:case} can be reliably detected by
our system.

Unlike other participating teams, 1-HKUST had very limited computing budget and resources
in our training and experiments:
one 24-core server PC (Dell PowerEdge R720 2 x 12C CPU, 128GB RDIMM memory and one NVIDIA GRID K1 GPU), and
one 6-core PC (Dell Alienware Aurora \@ 4.1Ghz, 6C CPU, 32GB DDR2 memory and one NVIDIA GeForce GTX 690 GPU).
Due to limited computing resources, the parameter tuning might not have been optimized, and
we strongly believe that our framework could achieve a better mAP rating if more
computing resources available and careful optimization tuning.

\begin{table}[t]
\begin{center}
\begin{tabular}{|c||c|c|c|c|c|c|c|c|}
\hline
Team name        & Number of object & mAP  \\
& categories won & \\
\hline
NUS              &       106                 &          0.372  \\
\hline
MSRA             &       45                        &    0.351  \\
\hline
UvA-Euvision&       21                        &    0.320  \\
\hline
\textbf{1-HKUST} &     \textbf{ 18   }                    &   \textbf{0.289}  \\
\hline
Southeast-CASIA  &      4                       &   0.304  \\
\hline
CASIA-CRIPAC-2   &      0                         &   0.286 \\
\hline
\end{tabular}
\end{center}
\caption{Number of object categories won without extra training data.}
\label{tab:result1}
\end{table}
\begin{table}[t]
\begin{center}
\begin{tabular}{|c||c|c|c|c|c|c|c|c|}
\hline
Team name        & Number of object \\
& categories won \\
\hline
GoogLeNet              &       138                   \\
\hline
CUHK-DeepID-Net             &       28                       \\
\hline
Deep-Insight &       27                          \\
\hline
\textbf{1-HKUST} (run 2) &     \textbf{ 3}                          \\
\hline
Berkeley-Vision &     1                          \\
\hline
NUS  &      1                         \\
\hline
UvA-Euvision   &      1                          \\
\hline
MSRA-Visual-Computing  &      0                         \\
\hline
MPG-UT   &      0                         \\
\hline
ORANGE-BUPT   &      0                         \\
\hline
Trimps-Soushen &      0                         \\
\hline
MIL &      0                         \\
\hline
 Southeast-CASIA &      0                         \\
\hline
CASIA-CRIPAC-2 &      0                         \\
\hline
\end{tabular}
\end{center}
\caption{Number of object categories won with and without extra
training data.  1-HKUST did not use extra training data.  }\label{tab:result2}
\end{table}

\bibliographystyle{ieee}
\bibliography{egbib}

\end{document}